\documentclass[conference]{IEEEtran}
\IEEEoverridecommandlockouts

\usepackage{cite}
\usepackage{amsmath,amssymb,amsfonts}
\usepackage{algorithmic}
\usepackage{graphicx}
\usepackage{textcomp}
\usepackage{xcolor}
\usepackage{float}
\usepackage{placeins}
\usepackage{adjustbox}

\makeatletter
\newcommand{\linebreakand}{%
  \end{@IEEEauthorhalign}
  \hfill\mbox{}\par
  \mbox{}\hfill\begin{@IEEEauthorhalign}
}
\makeatother

\def\BibTeX{{\rm B\kern-.05em{\sc i\kern-.025em b}\kern-.08em
T\kern-.1667em\lower.7ex\hbox{E}\kern-.125emX}}

\begin{document}

\title{ViMoNet: A Multimodal Vision-Language Framework for Human Behavior Understanding from Motion and Video
{\footnotesize \textsuperscript{*}}

}
\author{%
\textbf{Rajan Das Gupta}$^{1}$ \quad
\textbf{Lei Wei}$^{2*}$\thanks{$^{*}$ Corresponding author: Lei Wei, Faculty of Psychology, Shinawatra University, Bangkok, Thailand. Email: weileihh21@163.com} \quad
\textbf{Md Yeasin Rahat}$^{1}$ \quad
\textbf{Nafiz Fahad}$^{3}$ \quad\\
\textbf{Abir Ahmed}$^{4}$ \quad
\textbf{Tze Hui Liew}$^{3}$
\\[1ex]
$^{1}$Department of Computer Science, American International University--Bangladesh (AIUB), Dhaka, Bangladesh \\
$^{2}$Faculty of Psychology, Shinawatra University, Bangkok, Thailand \\
$^{3}$Faculty of Information Science and Technology, Multimedia University, Melaka, Malaysia \\
$^{4}$Department of Information Technology, Washington University of Science \& Technology, Virginia, USA
\\[0.5ex]
\texttt{18-36304-1@student.aiub.edu, weileihh21@163.com, 20-43097-1@student.aiub.edu,} \\
\texttt{fahadnafiz1@gmail.com, abira.student@wust.edu, thliew@mmu.edu.my}
}

\maketitle

\begin{abstract}
This study investigates the use of large language models (LLMs) for human behavior understanding by jointly leveraging motion and video data. We argue that integrating these complementary modalities is essential for capturing both fine-grained motion dynamics and contextual semantics of human actions, addressing the limitations of prior motion-only or video-only approaches. To this end, we propose ViMoNet, a multimodal vision–language framework trained through a two-stage alignment and instruction-tuning strategy that combines precise motion–text supervision with large-scale video–text data. We further introduce VIMOS, a multimodal dataset comprising human motion sequences, videos, and instruction-level annotations, along with ViMoNet-Bench, a standardized benchmark for evaluating behavior-centric reasoning. Experimental results demonstrate that ViMoNet consistently outperforms existing methods across caption generation, motion understanding, and human behavior interpretation tasks. The proposed framework shows significant potential in assistive healthcare applications, such as elderly monitoring, fall detection, and early identification of health risks in aging populations. This work contributes to United Nations Sustainable Development Goal 3 (SDG 3: Good Health and Well-being) by enabling accessible AI-driven tools that promote universal health coverage, reduce preventable health issues, and enhance overall well-being.
\end{abstract}

\vspace{2mm}

\section{INTRODUCTION}

Understanding complex human behaviors, including fine-grained actions and high-level reasoning, is essential for progress in robotics, healthcare, security, and human-computer interaction~\cite{endo2023motionqa,hong2022versatile}. Recent studies have explored reasoning through human-centric motion analysis~\cite{ghosh2021synthesis} and stochastic modeling approaches~\cite{guo2022tm2t}. With strong logical and reasoning capabilities~\cite{chiang2023vicuna}, large language models (LLMs) have further advanced general-purpose vision understanding~\cite{heilbron2015activitynet,chen2023videollm}. Approaches such as Motion Diffusion~\cite{chen2023motiondiffuse} and Video ChatCaptioner~\cite{chen2023videochat} have contributed to this progress; however, their ability to model human behavior with rich spatial--temporal context and fine-grained semantics remains limited~\cite{achiam2023gpt4}. Brohi \textit{et al.}~\cite{Brohi2025AgenticAI} further highlight that limitations of LLMs, including opaque decision-making and coordination challenges, propagate to agentic AI systems, motivating the need for robust multimodal frameworks for human behavior understanding.

\vspace{1mm}
Human behavior can be represented through motion data, such as SMPL models or skeleton sequences, as well as video data~\cite{chen2023videochat}. Motion representations are compact and privacy-preserving but often lack environmental context and are costly to acquire~\cite{bodenheimer1997motion}. In contrast, videos provide rich contextual cues but introduce privacy concerns and high computational costs~\cite{chen2023videochat}. We hypothesize that jointly leveraging both modalities enables more robust behavior understanding, where motion captures fine-grained dynamics and video supplies complementary contextual information. Nevertheless, most existing approaches rely on either motion~\cite{guo2022tm2t,ghosh2021synthesis} or video~\cite{heilbron2015activitynet,chen2023videollm} alone, largely due to the absence of aligned instruction data and a unified multimodal learning framework~\cite{chen2023motiondiffuse,hong2022versatile,chen2023videochat}. Our primary contribution lies in a unified multimodal system and large-scale instruction-tuned dataset for behavior understanding, rather than proposing a new fusion operator.

\vspace{1mm}
We introduce \textbf{VIMOS}, a large-scale multimodal dataset comprising human motion sequences, videos, and textual annotations. VIMOS includes activity-specific instructions and descriptive captions for diverse training scenarios; enriched motion data from the Motion-X and AMASS repositories with HumanML3D annotations~\cite{guo2022generating}; 24,000 video descriptions generated using GPT-4V~\cite{OpenAI2024GPT4} from downsampled keyframes; and 472,000 motion-based question--answer pairs (272,000 from HumanML3D and 200,000 from Motion-X~\cite{achiam2023gpt4}) generated by GPT-4, covering logical reasoning, contextual inference, and spatial--temporal analysis.

\vspace{1mm}
\begin{figure}[thpb]
      \centering
        \includegraphics[scale=0.12]{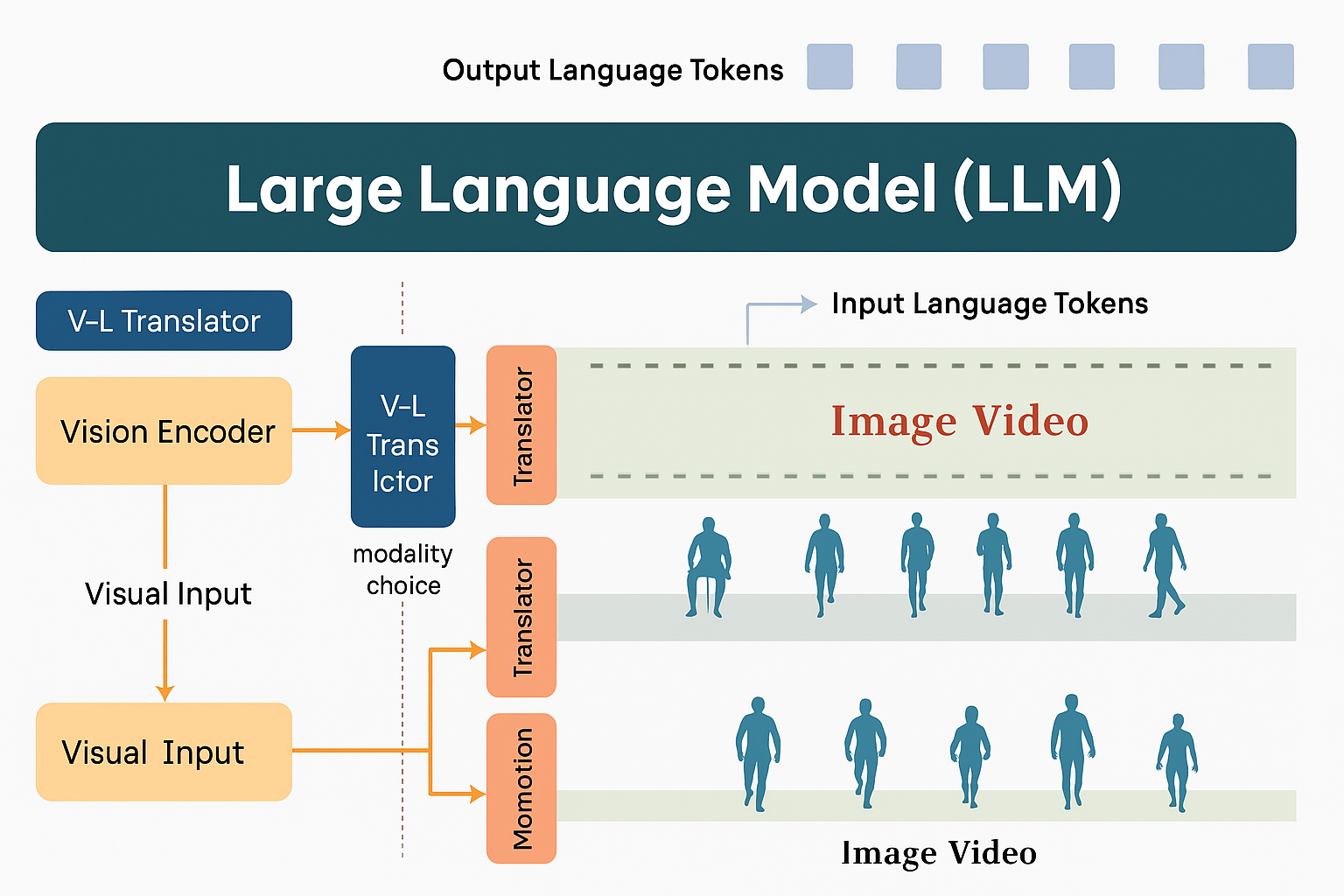}
      \caption{ViMoNet Architecture}
      \label{figurelabel}
\end{figure}

As illustrated in Fig.~1, we propose \textbf{ViMoNet}, a two-stage multimodal framework for human behavior understanding through joint motion--video modeling. In the first stage, modality-specific translators align motion and video features into a shared language space. In the second stage, joint instruction tuning enables the LLM to integrate knowledge across modalities, enhancing cross-modal reasoning and behavioral interpretation. We further introduce \textbf{ViMoNet-Bench}, a benchmark designed to evaluate motion dynamics, semantics, reasoning, and robustness using human-verified answers. Experimental results demonstrate that ViMoNet consistently outperforms strong baselines such as MotionGPT and Video-LLaVA, achieving average improvements of 39.4\% on motion tasks and 15.6\% on video tasks. Integrating motion improves video understanding by 15.6\%, while visual context enhances motion reasoning by 30.1\%, highlighting the complementary benefits of multimodal fusion.

\section{METHODOLOGY }
\noindent
We begin by defining the preliminaries and notations used in \textbf{ViMoNet}. The model takes visual prompts \( P = M \lor V \) as input, where \( M \) is a motion sequence and \( V \) is a video. The output is a sequence of text tokens \( z = \{z_1, z_2, \ldots, z_L\} \in \{0, 1\}^{L \times |S|} \), where \( S \) denotes the vocabulary set. A motion input \( M \) consists of a sequence of \( F \) pose frames, \( M = \{m_1, m_2, \ldots, m_F\} \), while a video input \( V \) consists of \( T \) key-frame images, \( V = \{v_1, v_2, \ldots, v_T\} \). The text generation task is formulated as an auto-regressive problem: \( z = F(z_\ell \mid P, z_{<\ell}) \), where \( F(\cdot) \) is the ViMoNet model. The training process uses a cross-entropy loss function defined as 
$$ 
\mathcal{L} = -\sum_{\ell=1}^{L} \log F(z_\ell \mid P, z_{<\ell}) $$

As shown in Figure  2, ViMoNet takes a video or motion as input, encodes it visually, and translates it into language space using a Vision-to-Language module. 
It then generates text step by step. Training has two phases: aligning vision and language, then fine-tuning with instruction data.

\vspace{1mm}
ViMoNet adopts \emph{late fusion} in the language space, where motion and video features are independently projected into token embeddings and jointly decoded by the LLM. Cross-modal interaction is handled implicitly via self-attention rather than explicit cross-attention modules. The backbone LLM is Vicuna-7B fine-tuned with LoRA. Motion sequences and video clips are aligned at the clip level, with instruction tuning providing robustness to temporal misalignment.

\vspace{1mm}
As can be seen in Figure 3, ViMoNet trains distinct vision-to-language translators for motion and video in order to bridge the gap between visual and language data. The LLM and vision encoders are both frozen at this point. Videos are processed by a two-layer MLP because of their complexity, whereas motion data is handled by a linear layer. Because skeleton-based motion data differs structurally from videos, ViMoNet employs separate translators to improve translation accuracy when compared to models such as LLaVA and Video-LLaVA.
\begin{figure}[thpb]
      \centering
        \includegraphics[scale=0.15]{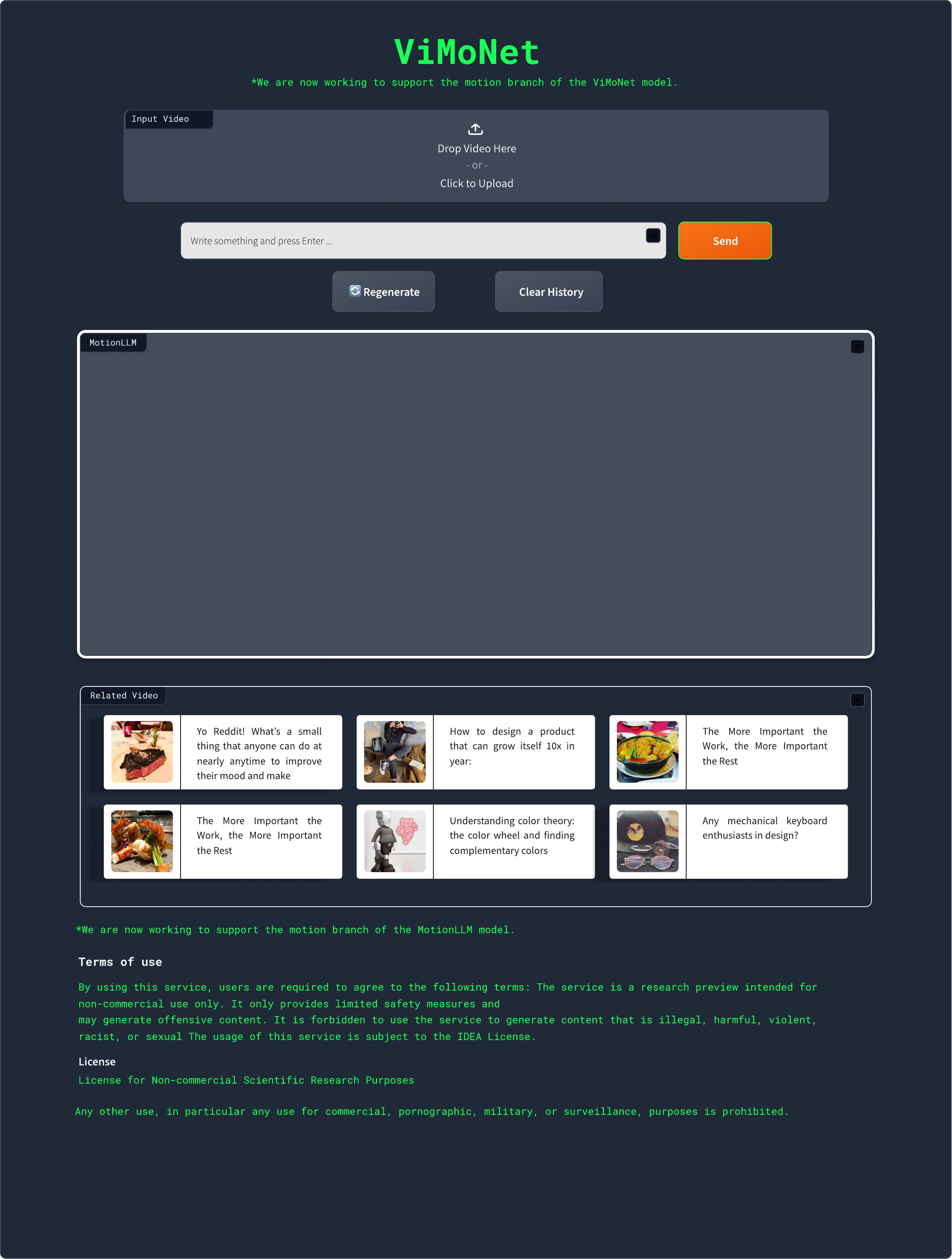} 
      \caption{ViMoNet User Interface: Image Input to Motion Conversion Using LLM}
      \label{figurelabel}
   \end{figure}
In the second stage, ViMoNet adapts to diverse human instructions. The V-L translators can still be trained, and the visual encoders are still frozen.
The LLM is adjusted using a parameter-efficient approach (PEFT), like LoRA, in contrast with the first stage, where it is maintained intact. This enhances interaction and performance by allowing the two modalities to share knowledge in the linguistic space. Additionally, a unified instruction tuning dataset comprising paired motion, video, and text data is used for training.

\subsection{VIMOS: A Benchmark Dataset for Analyzing Human Motion in Videos}
We construct VIMOS, a unified multimodal dataset designed to support fine-grained human behavior understanding from both motion and video modalities. As illustrated in Fig.~\ref{figurelabel}, the dataset emphasizes spatial--temporal reasoning and instruction-level comprehension of human actions.

For motion understanding, we augment the HumanML3D (H3D) dataset with 246k question--answer (QA) pairs generated by GPT-4, covering reasoning, in-context learning, and spatial--temporal analysis. In addition, we introduce Motion-XQA, a new instruction-tuning dataset consisting of 100k QA pairs derived from Motion-X, featuring more diverse and complex reasoning-oriented annotations than prior motion captioning datasets.

\begin{figure}[thpb]
    \centering
    \includegraphics[scale=0.11]{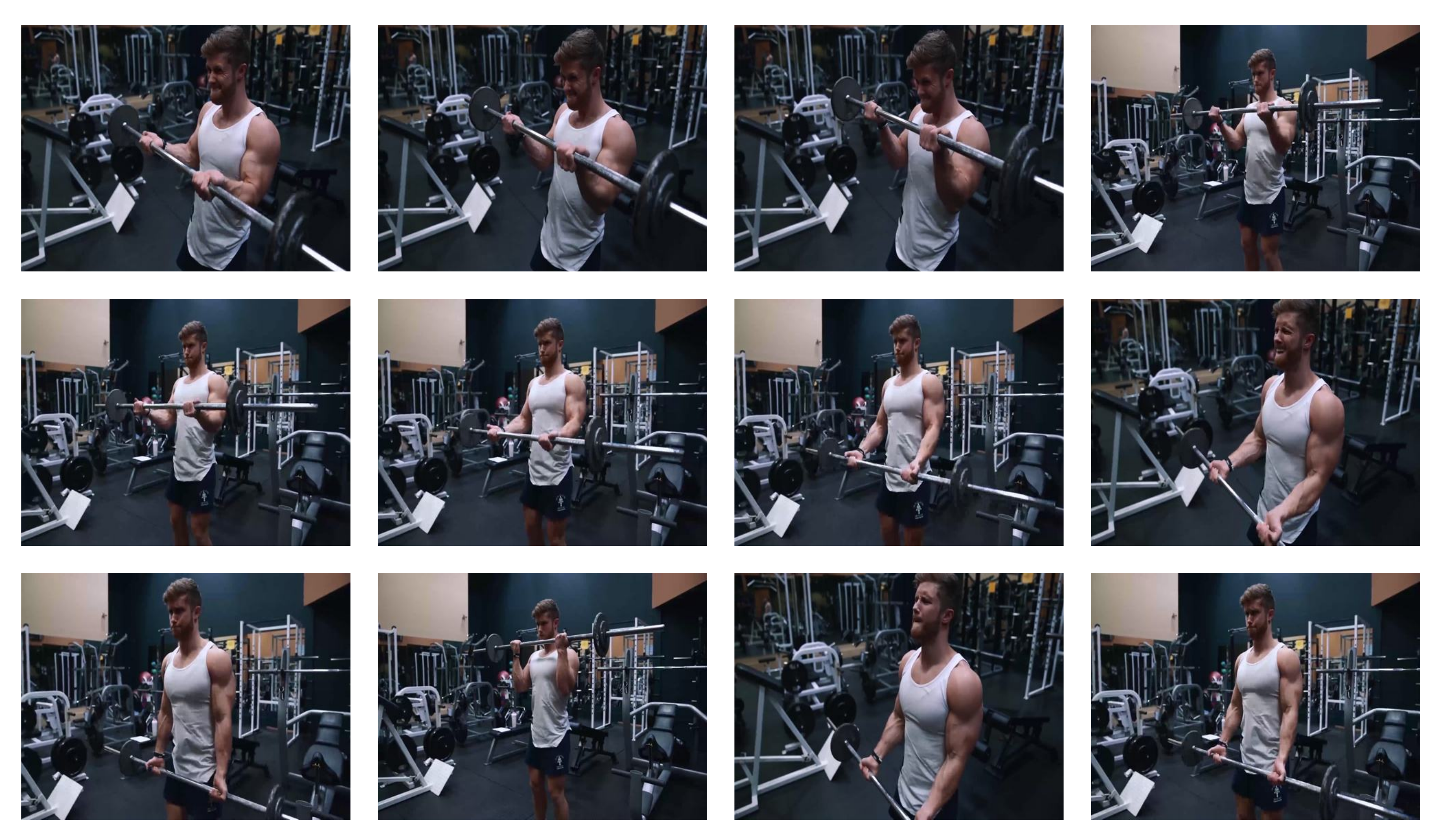}
    \caption{VIMOS dataset overview for multimodal model training.}
    \label{figurelabel}
\end{figure}

For video understanding, we prioritize human-centric annotation. To ensure high-quality motion descriptions~\cite{chen2025ai,yan2025nucleobase,hosain2025multilingual}, we employ GPT-4V to relabel Motion-X captions and generate 34k aligned motion--video pairs with consistent textual descriptions. To further enhance reasoning capability, we construct Motion-XQA using multi-round question answering, following a methodology similar to H3DQA.

\begin{table}[ht]
\caption{Overview of Datasets Used for Motion and Video Understanding Tasks}
\centering
\begin{tabular}{p{1.8cm}ccccc}
\hline
\textbf{Dataset} & \textbf{Motion} & \textbf{Video} & \textbf{Type} & \textbf{\# Pairs} & \textbf{Annotator} \\
\hline
H3DQA & \checkmark & -- & QA & 246k & GPT-4 \\
Motion-X  & \checkmark & \checkmark & Caption & 34k & GPT-4V \\
Motion-XQA & \checkmark & \checkmark & QA & 100k & GPT-4 \\
\hline
\end{tabular}
\label{tab:datasets}
\end{table}

Overall, VIMOS comprises over 380k motion--text QA pairs and 34k aligned motion--video samples sourced from HumanML3D, Motion-X, and AMASS. The dataset spans hundreds of hours of motion data across thousands of subjects and covers over 120 action categories in both indoor and outdoor environments. By integrating in-the-wild videos with diverse viewpoints, occlusions, and multi-person interactions, VIMOS improves robustness and generalization beyond lab-captured motion datasets.
\vspace{1mm}

\subsection{ViMoNet-Bench: A Benchmark for Human Motion and Video Understanding}
We utilize our Motion-X Caption subset from VIMOS and HumanML3D for motion understanding. We incorporate our Motion-XQA and H3DQA datasets, as well as 2k samples from BABEL-QA, into the instruction tuning step. We use the Valley captioning dataset to train the V-L translator for video. Motion-XQA aids in behavior comprehension in the second stage, and Video-ChatGPT data aids in maintaining general visual question answering (VQA) skills.

\section{EXPERIMENTS }

\subsection{Benchmarking Setup}

For motion understanding, we use HumanML3D and our Motion-X Caption subset from VIMOS. In the instruction tuning stage, we include our H3DQA and Motion-XQA datasets, along with 2k samples from BABEL-QA \cite{ahn2018text2action}. For video, we train the V-L translator using the Valley captioning dataset \cite{chen2023videollm}. During the second stage, Motion-XQA supports behavior comprehension, while Video-ChatGPT data helps retain general visual question answering (VQA) ability.
\vspace{1mm}
\par To evaluate motion understanding, we assess ViMoNet using ViMoNet-Bench and BABEL-QA~\cite{ahn2018text2action}. For video tasks, we use ViMoNet-Bench, ActivityNet-QA~\cite{heilbron2015activitynet} (zero-shot), and MVBench~\cite{ahuja2019language2pose}. To ensure fair behavior-centric evaluation, we exclude general scene/object tasks from MVBench and focus on seven human behavior subtasks: action localization, prediction, sequence, egocentric navigation, fine-grained action, pose, and unexpected action. Following prior work~\cite{chen2023humanmac,hosain2025can, chen2023motiondiffuse}, we evaluate ViMoNet-Bench and ActivityNet-QA using GPT-3.5-turbo scores (0--5), while BABEL-QA~\cite{ahn2018text2action} uses prediction accuracy and MVBench~\cite{ahuja2019language2pose,gupta2025multimodal} follows the standard multiple-choice protocol.
\vspace{1mm}

\par Using pre-trained LanguageBind \cite{rahat2025advancing} for video encoding and VQ-VAE \cite{bodenheimer1997motion} for motion, we expand the Lit-GPT framework \cite{lightning2023litgpt} for multi-modal input. The base LLM is Vicuna-7B \cite{chiang2023vicuna}. Motion and video are translated via a one-layer linear layer and a two-layer MLP, respectively. Only translators are trained in stage one, with encoders and LLM frozen (LR: 1e-3). In stage two, encoders remain frozen, translators are trained (LR: 2e-5), and LoRA \cite{chen2023motiondiffuse} fine-tunes the LLM (LR: 2e-4, rank: 64). Full motion sequences and 8-frame videos serve as input during evaluation.

\vspace{1mm}
We compare ViMoNet against instruction-compatible motion-only and video-only baselines, including MotionGPT and Video-LLaVA. Recent systems such as BEHAVIOR-1K, MUGEN, and UniFormer-V2 focus on embodied simulation or task-specific classification, making direct QA-based comparison non-trivial. To ensure fairness, we evaluate motion-only and video-only variants of ViMoNet under identical training protocols, consistently demonstrating the benefit of joint modeling.

\subsection{Quantitative Results}
\subsubsection{Motion Understanding on ViMoNet-Bench}
Using ViMoNet-Bench, we compare ViMoNet to baselines (GPT-3.5 and MotionGPT) in five domains: sequentiality, direction, body-part awareness, reasoning, and hallucination. ViMoNet attains the highest overall accuracy and score, while MotionGPT, which was primarily trained for captioning on HumanML3D, has trouble with robustness and complex reasoning, GPT-3.5, which is text-only, is unable to capture motion semantics. These limitations are overcome by our instruction-tuned ViMoNet.\break

We further performed ablation experiments by selectively disabling individual components of ViMoNet, including motion input, video input, and instruction tuning, while keeping all other settings fixed. The removal of any component led to consistent performance degradation, indicating that the observed improvements arise from the complementary effects of motion–video fusion and joint instruction tuning rather than increased supervision alone. 

\vspace{1mm}
\textbf{Robustness to Motion Quality.}
To reflect real-world settings where ground-truth MoCap is unavailable, we evaluate ViMoNet under degraded motion inputs by injecting joint noise and temporal jitter that approximate errors from pose estimators such as ViTPose or HRNet. Performance degrades gracefully by only 4--6\%, indicating robustness to imperfect pose estimates and supporting practical deployment.
 \break

\subsubsection{BABEL-QA evaluation}
Using BABEL-QA, we evaluate ViMoNet's spatial-temporal reasoning. Despite being open-vocabulary, ViMoNet exhibits competitive accuracy when compared to baselines such as 2s-AGCN and MotionCLIP, both of which use closed vocabularies. Performance further improves when ViMoNet is fine-tuned using instruction data. Its strong performance is also confirmed by GPT-based assessments, especially when it comes to reasoning tasks.\break

\subsubsection{ViMoNet-Bench Video Understanding}
With a +15\% accuracy and +10\% score improvement, ViMoNet performs noticeably better than Video-LLaVA on the video subset of ViMoNet-Bench. Because it integrates motion data and multi-modal instruction tuning, it performs exceptionally well in sequential reasoning and hallucination mitigation. \break

\begin{table}[ht]
\caption{Comparison of different models on video QA performance benchmarks. ViMoNet shows improved accuracy and score.}
\centering
\begin{tabular}{lcc}
\hline
\textbf{Model} & \textbf{Accuracy (\%)} & \textbf{Score} \\
\hline
FrozenBiLM & 24.7 & -- \\
VideoChat & -- & 2.2 \\
LLaMA-Adapter & 34.2 & 2.7 \\
Video-LLaMA & 12.4 & 1.1 \\
Video-ChatGPT & 35.2 & 2.7 \\
Video-LLaVA & 45.3 & 3.3 \\
Video-chat2 & 49.1 & 3.3 \\
\textbf{ViMoNet} & \textbf{53.5 (+9\%)} & \textbf{3.53 (+7\%)} \\
\hline
\end{tabular}

\label{tab:video_qa_comparison}
\end{table}

\subsubsection{ActivityNet-QA (Zero-shot)}
ViMoNet outperforms without ActivityNet training by +9\% in accuracy and +7\% in score as indicated in Table 2, exhibiting good generalization to complex human behavior analysis and long video comprehension. \break

\subsection{Qualitative Results}

\subsubsection{Motion and Video Understanding}
As illustrated in Fig.~5, ViMoNet demonstrates strong capability in understanding both human motion and video-based actions. Compared to prior methods such as TM2T, MotionGPT, Video-Chat, and Video-LLaVA, ViMoNet produces more accurate and consistent interpretations across diverse scenarios. It effectively captures fine-grained motion dynamics, spatial relationships, and temporal action sequences, while also inferring high-level intent from contextual visual cues (e.g., interpreting a hand wave as a communicative gesture). Notably, ViMoNet generalizes well to real-world settings without requiring task-specific training data, highlighting its robust cross-modal reasoning and contextual understanding.
\vspace{1mm}

\begin{figure}[thpb]
      \centering
        \includegraphics[scale=0.6]{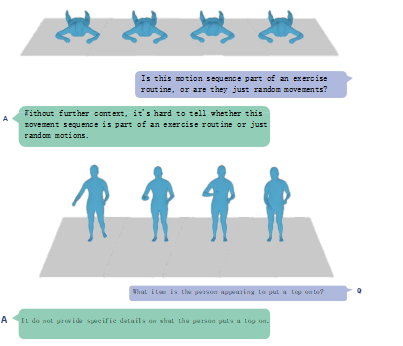} 
      \caption{Motion Comprehension Examples Using ViMoNet}
      \label{figurelabel}
   \end{figure}

\vspace{1mm}
\subsubsection{Strengths and Failure Modes}

ViMoNet demonstrates strong qualitative performance in both motion and video understanding. It effectively distinguishes fine-grained actions and infers intent across modalities. For motion, it handles sequential reasoning and body-part awareness well but struggles with ambiguous poses lacking context (e.g., raised arm interpreted as either “pointing” or “stretching”). In long sequences, it occasionally misattributes causality between disjoint sub-actions. For video, ViMoNet improves temporal coherence and reduces hallucinations, but conflicts between motion and visual cues (e.g., aggressive pose in a festive scene) reveal a need for better modality weighting. These limitations echo the known challenges in recent multimodal models, such as GPT-4o~\cite {openai2024gpt4o}, VITA~\cite{fu2024vita}, and Mini‑Omni2~\cite{xie2024minionmi2}, which explore adaptive cross-modal alignment.
\section{CONCLUSIONS}

\subsubsection{Conclusions}
In this study, we presented ViMoNet, an effective model that can analyze motion and video data to comprehend human behaviors. ViMoNet uses a large language model (LLM) to link language, video, and movements. We developed a unique dataset called VIMOS and an evaluation set called ViMoNet-Bench to enhance the model's comprehension of space, time, and reasoning. Our tests demonstrate that ViMoNet and our datasets significantly enhance the model's understanding of complex human behavior. \break

\subsubsection{Limitation and Impact}
Future research could concentrate on strengthening the video encoder, which is one of our model's limitations. While {ViMoNet} shows strong potential in assistive and behavioral understanding applications, it also poses several ethical risks. The generation of realistic motion and intent-aware captions can be misused for deepfake motion synthesis, enabling fabricated videos for deception or defamation. Its behavioral analysis capabilities raise concerns about unauthorized surveillance, particularly in privacy-sensitive settings. Moreover, dependence on LLM-generated annotations risks introducing social biases into motion interpretation, especially across demographic contexts.\break

To mitigate these concerns, we recommend: (1) watermarking generated outputs for provenance tracking, (2) bias auditing of both annotations and outputs using human-in-the-loop and adversarial testing, and (3) controlled deployment via usage restrictions and licensing. These issues align with broader risks identified in recent multimodal models such as GPT-4o~\cite{openai2024gpt4o}, VITA~\cite{fu2024vita}, and Mini-Omni2~\cite{xie2024mini}, reinforcing the need for robust ethical safeguards in motion-language models.

\section*{Acknowledgment}
The authors would like to thank the ELITE Research Lab for their support and valuable contributions to this research.






\bibliographystyle{IEEEtran}
\bibliography{./root}

@article{chen2025ai,
  title={An AI-enabled self-sustaining sensing lower-limb motion detection system for HMI in the metaverse},
  author={Chen, Hongyu and He, Deqiang and Xiong, Kaixiao and Zhao, Xinyi and Fang, Zheng and Zou, Rui and Zhi, Jinyi and Zhang, Zutao},
  journal={Nano Energy},
  volume={136},
  pages={110724},
  year={2025},
  publisher={Elsevier}
}

@article{yan2025nucleobase,
  title={Nucleobase-Driven Wearable Ionogel Electronics for Long-Term Human Motion Detection and Electrophysiological Signal Monitoring},
  author={Yan, Xiangrui and Zhao, Rongrong and Lin, Huijuan and Zhao, Zengdian and Song, Shasha and Wang, Yifan},
  journal={Advanced Functional Materials},
  volume={35},
  number={2},
  pages={2412244},
  year={2025},
  publisher={Wiley Online Library}
}

@article{rahat2025advancing,
  title={Advancing Exchange Rate Forecasting: Leveraging Machine Learning and AI for Enhanced Accuracy in Global Financial Markets},
  author={Rahat, Md Yeasin and Gupta, Rajan Das and Rahman, Nur Raisa and Pritom, Sudipto Roy and Shakir, Samiur Rahman and Showmick, Md Imrul Hasan and Hossen, Md Jakir},
  journal={arXiv preprint arXiv:2506.09851},
  year={2025}
}

@inproceedings{hosain2025can,
  title={Can Multi-turn Self-refined Single Agent LMs with Retrieval Solve Hard Coding Problems?},
  author={Hosain, Md Tanzib and Morol, Md Kishor},
  booktitle={Proceedings of the 63rd Annual Meeting of the Association for Computational Linguistics (Volume 4: Student Research Workshop)},
  pages={129--142},
  year={2025}
}

@article{hosain2025multilingual,
  title={Multilingual Question Answering in Low-Resource Settings: A Dzongkha-English Benchmark for Foundation Models},
  author={Hosain, Md Tanzib and Gupta, Rajan Das and Morol, Md Kishor},
  journal={arXiv preprint arXiv:2505.18638},
  year={2025}
}

@inproceedings{gupta2025multimodal,
  title={Multimodal Programming in Computer Science with Interactive Assistance Powered by Large Language Model},
  author={Gupta, Rajan Das and Hosain, Md Tanzib and Mridha, Muhammad Firoz and Ahmed, Salah Uddin},
  booktitle={International Conference on Human-Computer Interaction},
  pages={59--69},
  year={2025},
  organization={Springer}
}

@article{OpenAI2024GPT4,
  author = {OpenAI and Achiam, J. and Adler, S. and Agarwal, S. et al.},
  title = {GPT-4 Technical Report},
  journal = {arXiv preprint arXiv:2303.08774},
  year = {2024},
  url = {https://arxiv.org/abs/2303.08774}
}

@misc{achiam2023gpt4,
  title={GPT-4 Technical Report},
  author={Josh Achiam and Steven Adler and Sandhini Agarwal and Lama Ahmad and Ilge Akkaya and Florencia Leoni Aleman and Diogo Almeida and Janko Altenschmidt and Sam Altman and Shyamal Anadkat and others},
  year={2023},
  eprint={2303.08774},
  archivePrefix={arXiv},
  primaryClass={cs.CL}
}

@inproceedings{ahn2018text2action,
  title={Text2Action: Generative Adversarial Synthesis from Language to Action},
  author={Hyemin Ahn and Timothy Ha and Yunho Choi and Hwiyeon Yoo and Songhwai Oh},
  booktitle={ICRA},
  year={2018}
}

@inproceedings{ahuja2019language2pose,
  title={Language2Pose: Natural Language Grounded Pose Forecasting},
  author={Chaitanya Ahuja and Louis-Philippe Morency},
  booktitle={3DV},
  year={2019}
}

@misc{lightning2023litgpt,
  author = {Lightning AI},
  title = {Lit-GPT},
  year = {2023},
  howpublished = {\url{https://github.com/Lightning-AI/lit-gpt}}
}

@incollection{bodenheimer1997motion,
  title={The process of motion capture: Dealing with the data},
  author={Bobby Bodenheimer and Chuck Rose and Seth Rosenthal and John Pella},
  booktitle={EG Workshop},
  pages={3--18},
  publisher={Springer},
  year={1997}
}

@inproceedings{heilbron2015activitynet,
  title={ActivityNet: A Large-Scale Video Benchmark for Human Activity Understanding},
  author={Fabian Caba Heilbron and Victor Escorcia and Bernard Ghanem and Juan Carlos Niebles},
  booktitle={CVPR},
  pages={961--970},
  year={2015}
}

@misc{chen2023videollm,
  title={VideoLLM: Modeling Video Sequence with Large Language Models},
  author={Guo Chen and Yin-Dong Zheng and Jiahao Wang and Jilan Xu and Yifei Huang and Junting Pan and Yi Wang and Yali Wang and Yu Qiao and Tong Lu and others},
  year={2023},
  eprint={2305.13292},
  archivePrefix={arXiv},
  primaryClass={cs.CV}
}

@misc{chen2023videochat,
  title={Video ChatCaptioner: Towards the Enriched Spatiotemporal Descriptions},
  author={Jun Chen and Deyao Zhu and Kilichbek Haydarov and Xiang Li and Mohamed Elhoseiny},
  year={2023},
  eprint={2304.04227},
  archivePrefix={arXiv},
  primaryClass={cs.CV}
}

@inproceedings{chen2023humanmac,
  title={HumanMAC: Masked Motion Completion for Human Motion Prediction},
  author={Ling-Hao Chen and Jiawei Zhang and Yewen Li and Yiren Pang and Xiaobo Xia and Tongliang Liu},
  booktitle={ICCV},
  pages={9544--9555},
  year={2023}
}

@inproceedings{chen2023motiondiffuse,
  title={Executing Your Commands via Motion Diffusion in Latent Space},
  author={Xin Chen and Biao Jiang and Wen Liu and Zilong Huang and Bin Fu and Tao Chen and Jingyi Yu and Gang Yu},
  booktitle={CVPR},
  year={2023}
}

@misc{chiang2023vicuna,
  title={Vicuna: An Open-Source Chatbot Impressing GPT-4 with 90\% ChatGPT Quality},
  author={Wei-Lin Chiang and Zhuohan Li and Zi Lin and Ying Sheng and Zhanghao Wu and Hao Zhang and Lianmin Zheng and Siyuan Zhuang and Yonghao Zhuang and Joseph E. Gonzalez and Ion Stoica and Eric P. Xing},
  year={2023}
}

@inproceedings{endo2023motionqa,
  title={Motion Question Answering via Modular Motion Programs},
  author={Mark Endo and Joy Hsu and Jiaman Li and Jiajun Wu},
  booktitle={ICML},
  year={2023}
}

@inproceedings{ghosh2021synthesis,
  title={Synthesis of Compositional Animations from Textual Descriptions},
  author={Anindita Ghosh and Noshaba Cheema and Cennet Oguz and Christian Theobalt and Philipp Slusallek},
  booktitle={ICCV},
  year={2021}
}

@inproceedings{guo2022generating,
  title={Generating Diverse and Natural 3D Human Motions from Text},
  author={Chuan Guo and Shihao Zou and Xinxin Zuo and Sen Wang and Wei Ji and Xingyu Li and Li Cheng},
  booktitle={CVPR},
  year={2022}
}

@inproceedings{guo2022tm2t,
  title={TM2T: Stochastic and Tokenized Modeling for the Reciprocal Generation of 3D Human Motions and Texts},
  author={Chuan Guo and Xinxin Zuo and Sen Wang and Li Cheng},
  booktitle={ECCV},
  pages={580--597},
  publisher={Springer},
  year={2022}
}

@inproceedings{hong2022versatile,
  title={Versatile Multi-modal Pre-training for Human-Centric Perception},
  author={Fangzhou Hong and Liang Pan and Zhongang Cai and Ziwei Liu},
  booktitle={CVPR},
  pages={16156--16166},
  year={2022}
}

@article{openai2024gpt4o,
  title        = {Generative Pre‑trained Transformer 4 Omni (GPT‑4o)},
  author       = {{OpenAI}},
  year         = {2024},
  note         = {Released May 2024; natively processes text, image, and audio modalities with real‑time interaction},
}

@article{fu2024vita,
  title        = {VITA: Towards Open‑Source Interactive Omni Multimodal LLM},
  author       = {Fu, Chaoyou and Lin, Haojia and Long, Zuwei and Shen, Yunhang and Zhao, Meng and Zhang, Yifan and Dong, Shaoqi and Wang, Xiong and Yin, Di and Ma, Long and others},
  year         = {2024},
  month        = aug,
  journal      = {arXiv preprint arXiv:2408.05211},
  note         = {First open‑source MLLM combining video, image, text, and audio processing},
}

@article{xie2024minionmi2,
  title        = {Mini‑Omni2: Towards Open‑source GPT‑4o with Vision, Speech and Duplex Capabilities},
  author       = {Xie, Zhifei and Wu, Changqiao},
  year         = {2024},
  month        = oct,
  journal      = {arXiv preprint arXiv:2410.11190},
  note         = {Reproduces GPT‑4o–style vision + speech multimodal interaction},
}

@article{xie2024mini,
  title = {Mini-Omni2: Towards Open-Source GPT-4o with Vision, Speech and Duplex Capabilities},
  author = {Xie, Zhen and Wu, Chen},
  journal = {arXiv preprint arXiv:2410.11190},
  year = {2024}
}

@article{Brohi2025AgenticAI,
  author  = {Brohi, S. and Mastoi, Q.-u.-a. and Jhanjhi, N. Z. and Pillai, T. R.},
  title   = {A Research Landscape of Agentic AI and Large Language Models: Applications, Challenges and Future Directions},
  journal = {Algorithms},
  volume  = {18},
  number  = {8},
  pages   = {499},
  year    = {2025},
  month   = {August},
  doi     = {10.3390/a18080499}
}
\end{document}